# Background Image Generation Using Boolean Operations

Kardi Teknomo
Ateneo de Manila University
Quezon City, 1108 Philippines
+632-4266001 ext 5660

teknomo@gmail.com

Proceso Fernandez
Ateneo de Manila University
Quezon City, 1108 Philippines
+632-4266001 ext 5660

pfernandez@ateneo.edu

## ABSTRACT
Tracking moving objects from a video sequence requires segmentation of these objects from the background image. However, getting the actual background image automatically without object detection and using only the video is difficult. In this paper, we describe a novel algorithm that generates background from real world images without foreground detection. The algorithm assumes that the background image is shown in the majority of the video. Given this simple assumption, the method described in this paper is able to accurately generate, with high probability, the background image from a video using only a small number of binary operations.

## 1. INTRODUCTION
Tracking and modeling people from video sequence is a challenging task of recent research activities. Such people-tracking usually employs some form of background subtraction, one of the best tools for segmenting foreground from background images. Once foreground images have been extracted, then the desired algorithms (e.g., motion tracking, face recognition, etc.) using these images can be performed. Given a background image as reference, an image containing a foreground object over the same background is compared at every pixel position. The detected differences (passing some noise threshold) determine the pixels of the foreground image.

The problem with background subtraction is that it requires that the background image is already available. Unfortunately, this is not always the case. The background image has been traditionally searched manually or automatically from the video images when there are no objects, as suggested by Matsuyama [1] and Tsuchikawa [2] and more recently, automatic background generation had been suggested. For example, Haritaoglu et al [3] used gray level subtraction of Gaussian mixtures, while McKeena et al [4] used adaptive background subtraction method that combines color and gradient information. In those previous studies, object detection is necessary to obtain the background. In the case where many objects remain in the scene, the background cannot be updated using the traditional method.

Several other studies about background modeling do not use object detection anymore. Instead, they use techniques such as median filtering [5, 6], medoid filtering [7], approximated median filtering [8, 9], linear predictive filter [8], non-parametric model [9], Kalman filter [10-12] and adaptive smoothening [13]. However, most of these methods have high computational complexity.

In this paper, we propose a novel automatic background generation for people-tracking using simple Boolean operations without object detection. Our algorithm gets the background image even though some objects still remain in the scene. At least, three slices of image from the same sequence of video scene is needed to produce a background image, assuming that for every pixel position, the background occurs in the majority of the video. The algorithm is robust enough to handle even the cases when objects overlapping with each other at some frames, and not overlap on other frames. The advantage of our proposed system is the computational speed of only $O(R)$-time, which depends only on the resolution $R$ of an image and impressive accuracy (at least 99% accuracy in most cases satisfying our assumption) can be gained only within a manageable number of frames. It can be performed for both gray level and color video sequence.

The paper is organized as follows. Section 2 describes the terminologies and assumptions that we used for our method. The theory to generate the background and the derivation involved are described in Section 3 Section 4 provides several examples of the implementation of the algorithm. A set of conclusions in Section 5 completes the paper.

## 2. TERMINOLOGIES AND ASSUMPTIONS
We assume that the camera is stationary. If a video is taken from a fixed location with a certain focus, the images in each video slice can be separated into moving objects and a background. If the objects are defined as the things that move over time excluding noise, then the background image, or simply background, is an image of the environment, which does not change over time. We also assumed that the light of the environment changes only slowly relative to the motions of the people in the scene and the number of people does not occupy the scene for the most of the time at the same place. Since video of pedestrian traffic normally has the property that people on the scene are moving and the crowd only partially occludes the background scene, the algorithm is expected to work in this scenario.

The video is displayed at the normal rate of 25 to 30 frames per second. Each frame of the image sequence consists of a grid of pixels, and the number of pixels in a frame determines the frame's resolution. Each pixel in an image can be expressed as an intensity level. The intensity values are used for various image-related algorithms.

Generally, however, the algorithm will work whenever the following single important assumption holds: for each pixel position, the majority of the pixel values in the entire video contain the pixel value of the actual background image (at that position). Note that the entire background image does not even have to appear in any frame of the video, as long as each part of the background is shown in the majority of the video, then the algorithm is expected to work accurately.

We shall refer to the original image from the video slices as the original image. Image that contains only the object without the background is called an object image. The original image thus consists of the background and object images.

In background subtraction, the object image has a white value (zero) at spatial coordinate $(r, c)$ only if the intensity level difference between the original image and the background is small at the given coordinates, as determined by the threshold. The small difference that results from the segmentation is due to noise.

One of the traditional methods to generate a background image is by using the mode of intensity value in each pixel over several slices of image. Given $n$ images from the stationary camera, the mode of intensity level of all pixels in the same location $(r, c)$ is calculated and the result of mode-image may produce the background. Calculating the mode of all intensity values for all pixels over time, however, is rather time consuming, and requires large space. Another idea is needed to reduce the calculation time as presented in Section 3.

## 3. BACKGROUND IMAGE GENERATION
### 3.1 Boolean Operation to Determine Mode

Let $b(r, c, t)$ be the background pixel at location $(r, c)$ at time $t$ and let $\#b(r, c)$ be the number of background pixels at location $(r, c)$ for the whole image sequence of period $T$. Similarly, let $g(r, c, t)$ be the foreground pixel at location $(r, c)$ at time $t$ and let $\#g(r, c)$ be the number of foreground pixels at location $(r, c)$ for the whole image sequence of period $T$. Then, $\#b(r, c) + \#g(r, c) = T$.

Our assumption stated that for any pixel location $(r, c)$, $\#b(r, c) > \frac{1}{2}T$. This assumption normally holds in the video sequence scenes where people on the scene are moving and the crowd only partially occludes the background scene.

The simplest scenario happens when we have $T=3$ images with non-overlapping objects. Non-overlapping objects means at any pixel location $(r, c)$, at most $\#g(r, c)=1$ (thus, we have $\#b(r, c) \geq T - \#g(r, c) = 3-1 = 2$). What we obtain is actually the largest frequency of the pixel value (i.e. mode) at a pixel location $(r, c)$ for any location. For $T=3$, we need at least $\#b(r, c)/T=2/3$ or approximately 67% of the time sequence.

As we sample the video at say 25 frames per second, a few minutes of video sequence will provide us with a large value of $T$. Using only 3 images would probably not give accurate results in this case because $3/T$ would be negligibly small. However, if the basic assumption ($\#b(r, c) > \frac{1}{2}T$) is satisfied, then the algorithm is expected to work. In real time video sequence, we can set $T$ fixed at a large number to ensure that the background is denser than the foreground over time (e.g. to allow periods of time where only background are covered in the pixel location).

Doing that at pixel level, we can extend the information at image level using the same sequence of Boolean operations. However, at the image level, things are much simpler because we only need to make sure that the objects will not cover the background most of the time at the same place. When an object is moving, at certain pixel location, it will cover the background at some period but most of the time the moving object will not cover that background. When we go down further to the bit level of a pixel, the frequency of the value of the background is expected to be even bigger since the bit value of the foreground could match that of the background for a given bit position.

**Proposition 1**: For three frames of image sequence $x_1$, $x_2$, and $x_3$, the background image $B$ can be obtained using

$$B = x_3(x_1 \oplus x_2) + x_1 x_2 \qquad (1)$$

Where $\oplus$ is the exclusive disjunctive bit operator.

**Proof.**
Consider three binary images, $x_1$, $x_2$, and $x_3$. The object is represented by 0 and the background is represented by 1. The eight possibilities are shown in Table 1.

Table 1. **Eight Possibilities of Three Binary Images and Its Boolean function**

| | | | | | | | | |
|---|---|---|---|---|---|---|---|---|
| $x_1$ | 0 | 0 | 0 | 0 | 1 | 1 | 1 | 1 |
| $x_2$ | 0 | 0 | 1 | 1 | 0 | 0 | 1 | 1 |
| $x_3$ | 0 | 1 | 0 | 1 | 0 | 1 | 0 | 1 |
| S | 0 | 0 | 0 | 1 | 0 | 1 | 1 | 1 |

The Boolean mode function $S$ of the table happens when the number of one (1) entries is larger than half of the number of images. Mathematically, we can write a Boolean function $S$ defined as the mode value defined as

$$S = \begin{cases} 1 & if \quad \sum_{i=1}^{n} x_i \geq \lceil \frac{n}{2}+1 \rceil, \quad n \geq 3 \\ 0 & otherwise \end{cases} \qquad (2)$$

Where $\lceil .. \rceil$ is the ceiling function and $n$ is the number of images in the image sequence. For 3 images, the background image $B$ can be taken as the value of 1 in $S$, or

$$B = \bar{x}_1 x_2 x_3 + x_1 \bar{x}_2 x_3 + x_1 x_2 \bar{x}_3 + x_1 x_2 x_3 \qquad (3)$$

By the definition of the exclusive disjunctive operator (XOR), i.e., $x_1 \oplus x_2 = \overline{x}_1 x_2 + x_1 \overline{x}_2$, equation (3) can be rearranged into

$$B = \left(\overline{x}_1 x_2 + x_1 \overline{x}_2\right) x_3 + x_1 x_2 \left(\overline{x}_3 + x_3\right),$$

which can be simplified into equation (1). Gray scale images and color intensities can be converted to equivalent binary numbers for the execution of the Boolean operation explained above. The result of the operation can then be converted back to gray or color scale intensities. Thus, the Boolean operation in equation (1) can be used for both gray and color images as represented by RGB intensity values. QED

To show that the Boolean operation indeed produces the background image at gray level and color images, we set a very simple experiment as shown in Figure 1. First, we make a ramp background, then we put object image at RGB value of $g_1$ = [0, 0, 64], $g_2$ = [0, 0, 128] and $g_3$ = [0, 0, 255] in the three consecutive images in such away so that the objects do not overlap with each other. This assures us that for every pixel position in the sequence, the background image pixel value is in the majority of the image sequence. Then we can use equation (1) which, translated into simple code of Boolean operation, is

$$b = \left(g_3 \text{ and } \left(g_1 \text{ xor } g_2\right)\right) \text{ or } \left(g_1 \text{ and } g_2\right) \quad (4)$$

Since the operation can be performed either at the image level or at the pixel level, the implementation of three images used only four binary operations. For RGB image, the computation is given at each color as

*Red_Background* = ($R_3$ and ($R_1$ xor $R_2$)) or ($R_1$ and $R_2$)

*Green_Background* = (g3 And (g1 Xor g2)) Or (g1 And g2)

*Blue_Background* = (b3 And (b1 Xor b2)) Or (b1 And b2)

Where $r_i$, $g_i$, $b_i$ are red, blue and green intensity level of image $i$.

The value of $g_3$ and $\left(g_1 \text{ xor } g_2\right)$ are zero on the background and a ramp on the objects, while $g_1$ and $g_2$ give original ramp on the background and black (0, 0, 0) at bottom-left object and (0, 0, 64) for top-right object. When we combine the last two images using Boolean operator *Or*, we get back exactly the background image. The operation produces back perfect background because we do not put any noise. This process to obtain back the background is shown in Figure 2.

### 3.2 Background Generation Algorithm

To obtain the general formulation of the background image generation, we propose a multi level mode. At the first level, for each background image, three frames are selected at random from the image sequence to produce a background image using equation (1). The three background images produced at the first level are combined using equation (1) to yield better background image at the second level. The procedure is repeated until desired level L.

Let $img_1$, $img_2$, $img_3$ be 3 images (frames) selected at random, and with replacement (i.e., an image selected can be selected again). Define a getImageMode($img_1$, $img_2$, $img_3$) as a function that computes the modal image using the following algorithm

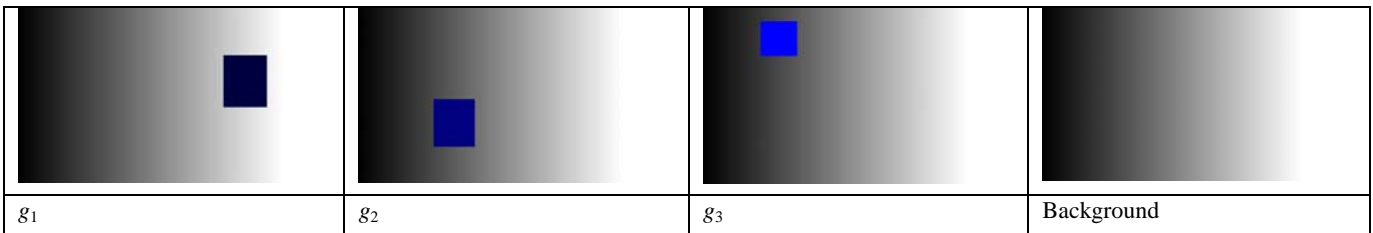

| $g_1$ | $g_2$ | $g_3$ | Background |

**Figure 1. Simple experimental images with one non-overlapping object on each image added from the most right background image.**

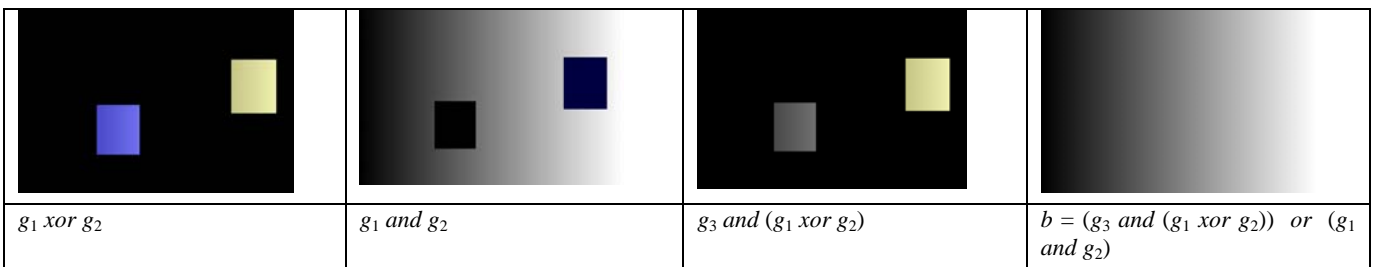

| $g_1$ xor $g_2$ | $g_1$ and $g_2$ | $g_3$ and ($g_1$ xor $g_2$) | $b = (g_3$ and $(g_1$ xor $g_2)) $ or $ (g_1$ and $g_2)$ |

**Figure 2. Computational process of the simple experimental images. The last image produces back the background image.**

```
Algorithm getImageMode(img1, img2, img3)
Input: 3 r x c images img1, img2 and img3
Output: modal image
1  for row ← 1 to r
2      for col ← 1 to c
3          p1 ← pixel at (row, col) for img1
4          p2 ← pixel at (row, col) for img2
5          p3 ← pixel at (row, col) for img3
6          p ← getPixelMode(p1, p2, p3)
7          let pixel at (row, col) for
                  modal_image be p
8  return modal_image
```

The Background Generation algorithm requires the input video *V* which is sequence of frames, and *L* as the desired level. The details of this algorithm are given below.

```
Algorithm BackgroundGeneration(V, L)
Input: video V and level L > 0
Output: background image
1  resultArray ← [] // an empty array of images
   // first, compute level 1 results
2  for i ← 0 to 3^(L - 1) - 1
3      img1 ← frame randomly selected from V
4      img2 ← frame randomly selected from V
5      img3 ← frame randomly selected from V
6      resultArray[i] ← getImageMode(img1,
                                    img2, img3)
   // next, compute results for other levels
7  for i ← 2 to L
8      for j ← 0 to 3^(L - i) - 1
9          img1 ← resultArray[3j]
10         img2 ← resultArray[3j+1]
11         img3 ← resultArray[3j+2]
12         resultArray[j] ←
               getImageMode(img1, img2, img3)
13 return resultArray[0] // modal image at level
   L
```

Function `getPixelMode(p1, p2, p3)` and `getImageMode(img1, img2, img3)` are computed using equation (1) at pixel and image level.

## 4. RESULTS AND ANALYSIS

### 4.1 Sample Results

Some sample test videos were used to demonstrate the algorithm. Figure 3 shows the results from an input video with only 2 pedestrians walking in a tiled corridor. The short video consists of 304 frames running at 20 hz. Because the background image pixels appear in an overwhelming majority of the frames, the background is generated even after just 1 level of processing. Note that video was used as-is, i.e., without any pre-processing. Only the moving objects disappear while non-moving objects are considered as the background. Similarly, the output after applying the algorithm did not require any post-processing. Note further that the result was generated using 1 sample run only. It is possible to generate a background image that (incorrectly) includes the 2 pedestrians in the image. However, as will be proven in the next section, the probability of such incorrect result is small, and this probability is reduced very quickly as more levels of processing are done.

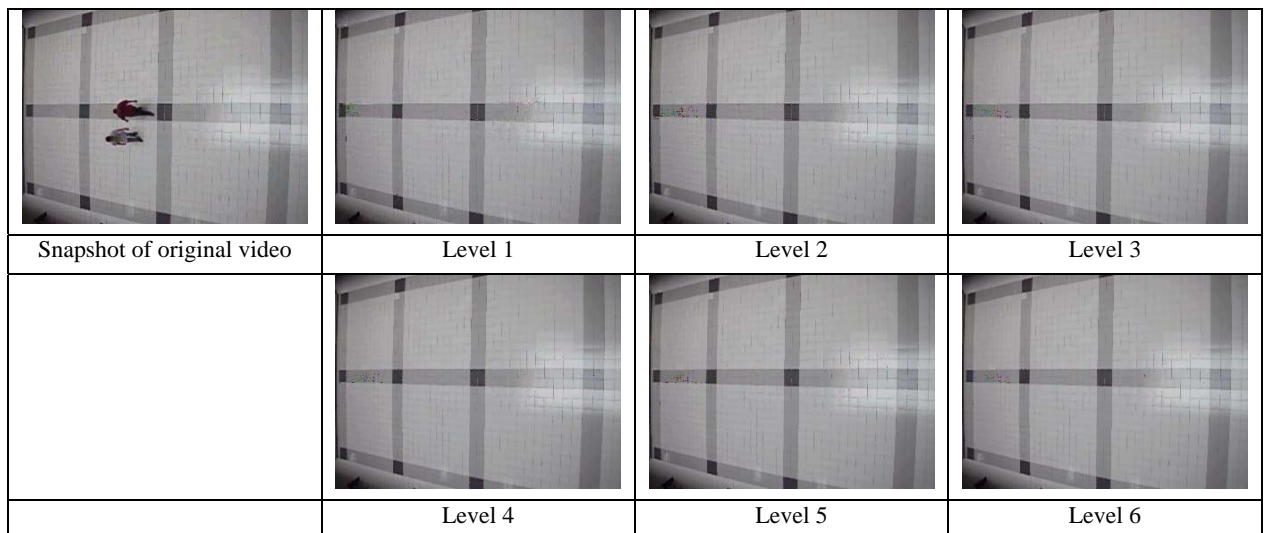

**Figure 3. Application of the algorithm on a video of 2 pedestrians on a corridor**

The second example, shown in Figure 4, illustrates the reliable performance of the algorithm even for complex videos. The complexity comes from the strip background, similar color between the moving objects and the background and crowded pedestrians that occupied the scene. Taken from the 6th floor building in Sendai Japan during a Tanabata Festival, the video was used in Teknomo's dissertation and has been used by several other papers. This video contains 256 frames at 2hz showing pedestrians crossing the street. Even though no single frame contains the entire background image, the background image was accurately generated after only 3 levels of processing. The succeeding level does not improve much of the accuracy.

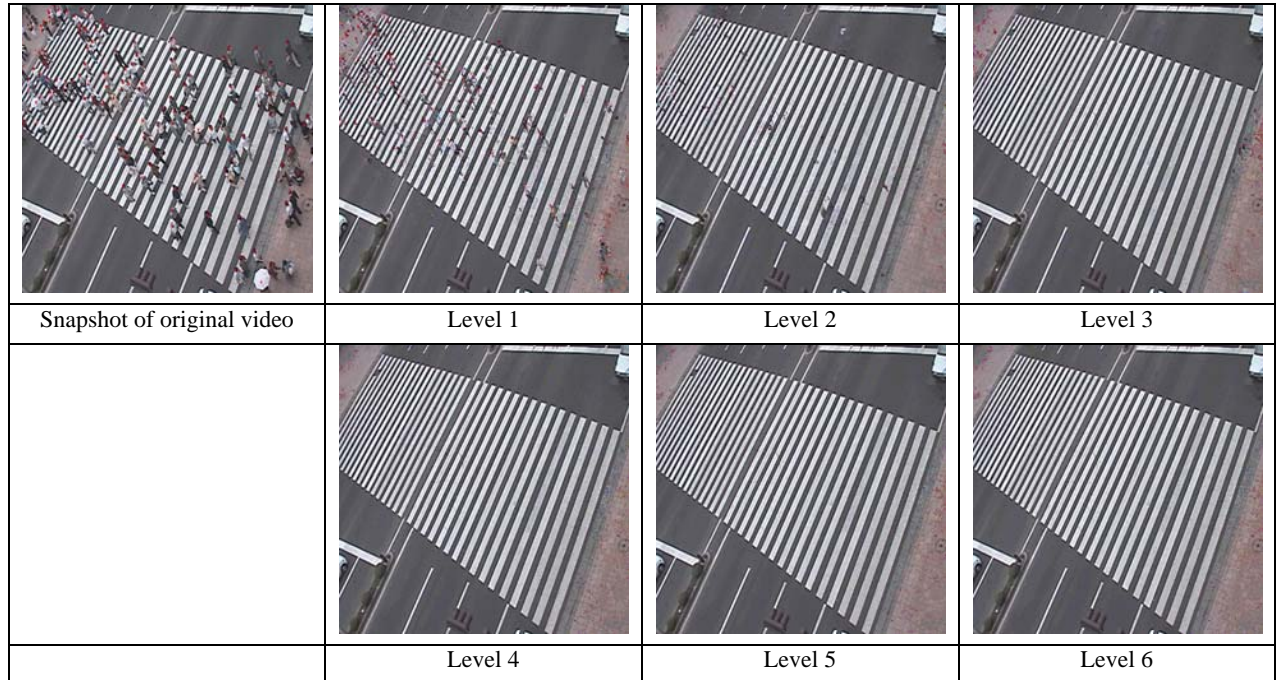

**Figure 4. Application of the algorithm on a video of many pedestrians crossing a street**

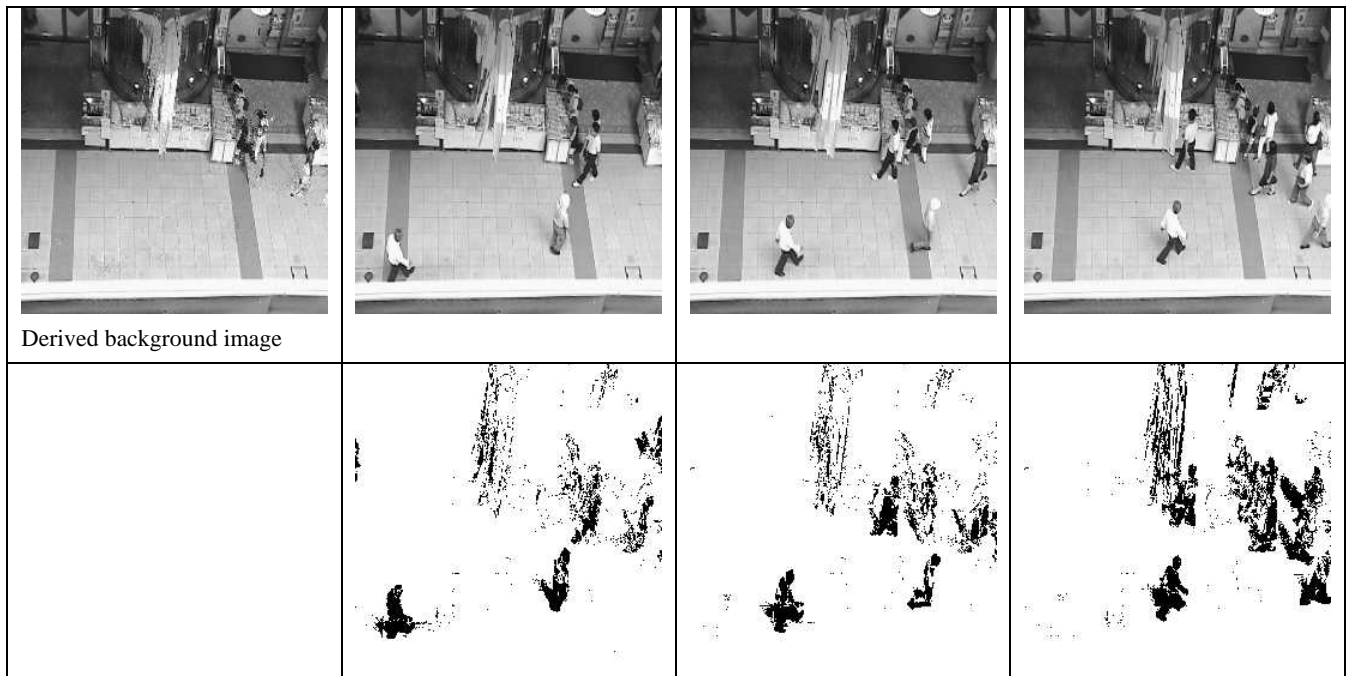

**Figure 5. Application of background subtraction on the derived background image.**

Finally Figure 5 illustrates how the background subtraction is done. Using the derived background image (top left) as basis, foreground images are computed by detecting differences between the image and the background. The figures show the original snapshot images in the topside and the result from binarized background subtraction are presented in lower side**.** The papers and shop display in the top center of image are not moving so they can be considered as background.

## 4.2 Expected Accuracy

Given a video that has $F$ frames, suppose that at any given pixel position ($r$, $c$) the modal pixel value has frequency $m \in (0.5F, F]$. This implies that at every bit position $i$ of the pixel integer value, the frequency that the modal bit value (0 or 1) for the given pixel across all the frames is $m$ or greater.

Consider a fixed bit position $i$ on some fixed pixel position ($r$, $c$). For a frame selected randomly from the entire video sequence, the probability $p_0$ that the $i$-th bit at pixel ($r$, $c$) of the frame is the actual modal bit value for that given position is therefore given by $p_0 \geq m/F$.

Consequently, for 3 frames selected at random and with repetition (i.e., a previously selected frame can be selected again), the probability that the modal bit for the 3 frames is also the modal bit for the entire video sequence is given by

$$p_1 = p_0^3 + 3(p_0)^2(1 - p_0). \qquad (5)$$

The first addend in the above formula gives the probability that the modal bit occurs in all of the 3 frames, while the second addend describes the scenario that exactly 2 of the 3 frames contain the modal bit at the considered position.

A second-level modal bit determination, based on the results of the first-level modal bit computations will yield the following probability of accurate prediction:

$$p_2 = p_1^3 + 3(p_1)^2(1 - p_1). \qquad (6)$$

A recursive formula can be easily observed from the preceding derivations. In general, at level $l$, the probability $p_l$ that the modal bit predicted is the actual modal bit can be computed by

$$p_l = (p_{l-1})^3 + 3(p_{l-1})^2(1 - p_{l-1}). \qquad (7)$$

The table 2 gives the computed values using some specific initial probabilities and across several levels.

The table shows, in particular, that even if the modal bit at the considered position occurs at a low 60% of the frames, the probability of accurate modal bit determination is already more than 99% at 6 levels. The number of frames that must be randomly selected for this is exactly $3^6 = 729$, or roughly 27 seconds of video data.

On the other hand, if the modal bit occurs more frequently at, say, 80%, then only 3 levels are needed to achieve at least 99% accuracy. These 3 levels require only $3^3 = 27$ frames, or roughly 1 second of video data!

**Table 2 Probabilities of accurate modal bit prediction for different levels**

| $p_0$ | $p_1$ | $p_2$ | $p_3$ | $p_4$ | $p_5$ | $p_6$ |
|---|---|---|---|---|---|---|
| 0.50 | 0.50 | 0.50 | 0.50 | 0.50 | 0.50 | 0.50 |
| 0.55 | 0.575 | 0.611 | 0.664 | 0.737 | 0.829 | 0.923 |
| 0.60 | 0.648 | 0.716 | 0.803 | 0.899 | 0.972 | 0.998 |
| 0.65 | 0.718 | 0.807 | 0.902 | 0.973 | 0.998 | 1.000 |
| 0.70 | 0.784 | 0.880 | 0.960 | 0.995 | 1.000 | 1.000 |
| 0.75 | 0.844 | 0.934 | 0.988 | 1.000 | 1.000 | 1.000 |
| 0.80 | 0.896 | 0.970 | 0.997 | 1.000 | 1.000 | 1.000 |
| 0.85 | 0.939 | 0.989 | 1.000 | 1.000 | 1.000 | 1.000 |
| 0.90 | 0.972 | 0.998 | 1.000 | 1.000 | 1.000 | 1.000 |
| 0.95 | 0.993 | 1.000 | 1.000 | 1.000 | 1.000 | 1.000 |
| 1 | 1 | 1 | 1 | 1 | 1 | 1 |

## 4.3 Space and Time Complexity

The space requirement for the algorithm is dependent on the resolution $R$ of the image, the number $F$ of frames in the video and the desired number $L$ of levels. In particular, the space complexity is given by the function $O(RF + R3^L)$. Although there is an exponential function involving $L$, the fact that $L$ will probably not exceed 6 reduces the space complexity to simply $O(RF)$, which is the same complexity as that of the input video. In fact, when the user wants exactly 6 levels of computations, then the required space for the results array is approximately equal to about 27 seconds of video clip only.

The time complexity is also impressive. The basis of the computation is that of the modal bit for 3 bits, and this can be done in $O(1)$-time. This implies that the getPixelMode function, which computes the modal bit for each of the bit for a fixed-length pixel value, can also be done in $O(1)$-time. Therefore, the computation of the resulting image from 3 given images can be done in $O(R)$-time, where $R$ is the resolution of the image. Finally, the number of images to be processed in $L$ levels is given by $O(3^L)$. However, since $L \leq 6$, then this is actually $O(1)$. Consequently, the entire algorithm runs in $O(R)$-time only, which means it depends only on the resolution of an image.

## 5. CONCLUSION

A new algorithm that can generate the background from real world videos was described in this paper. The algorithm does not need to detect the existence of the objects and is able to obtain the background even though some objects remain in the scene. It was further shown that, say every part of the background appears in at least 60% of the video, and then algorithm will be able to extract the background with over 99% accuracy after applying only a few levels of processing. What is further astonishing is the fact that this novel algorithm has very low time and space complexities.

Future work can focus on applying additional steps for the algorithm, including some pre-processing (e.g., histogram equalization, low pass filters and image adjustments) and post-processing (e.g., morphological image processing and applying threshold values) steps.